\documentclass[letterpaper]{article} 
\usepackage[]{aaai23}  
\usepackage{times}  
\usepackage{helvet}  
\usepackage{courier}  
\usepackage[hyphens]{url}  
\usepackage{graphicx} 
\usepackage{natbib}  
\usepackage{caption} 
\usepackage{algorithm}
\usepackage{algorithmic}

\usepackage{newfloat}
\usepackage{listings}
\DeclareCaptionStyle{ruled}{labelfont=normalfont,labelsep=colon,strut=off} 
\floatstyle{ruled}
\newfloat{listing}{tb}{lst}{}
\floatname{listing}{Listing}
\usepackage{url}
\usepackage{booktabs}
\usepackage{multirow}
\usepackage{amssymb}
\usepackage{pifont}
\usepackage{tabularx}
\usepackage{adjustbox}
\usepackage{amsmath,amsfonts,bm}
\usepackage{lipsum}
\usepackage{comment}
\usepackage{subcaption}
\usepackage{diagbox}
\usepackage{cleveref}
\usepackage{longtable}
\usepackage{xcolor}

\newcommand{\ours}{\textsc{LIQUID}}
\newcommand{\bioasq}{BioASQ}

\newcommand{\quoref}{Quoref}

\newcommand{\multispanqa}{MultiSpanQA}

\newcommand{\bertbase}{BERT$_\texttt{base}$}
\newcommand{\roberta}{RoBERTa}
\newcommand{\robertabase}{RoBERTa$_\texttt{base}$}
\newcommand{\robertalarge}{RoBERTa$_\texttt{large}$}

\newcommand{\coreflarge}{CorefRoBERTa$_\texttt{large}$}
\newcommand{\biobert}{BioBERT}
\newcommand{\biobertbase}{BioBERT$_\texttt{base}$}
\newcommand{\biobertlarge}{BioBERT$_\texttt{large}$}

\title{\ours: A Framework for List Question Answering Dataset Generation}

\author{
    {Seongyun Lee},\equalcontrib\textsuperscript{\rm 1}
    {Hyunjae Kim},\equalcontrib\textsuperscript{\rm 1} 
    {Jaewoo Kang}\textsuperscript{\rm 1,2}
}
\affiliations{
    \textsuperscript{\rm 1}Korea University,
    \textsuperscript{\rm 2}AIGEN Sciences

    \{sy-lee, hyunjae-kim, kangj\}@korea.ac.kr
}

\begin{document}

\maketitle

\begin{abstract}
Question answering (QA) models often rely on large-scale training datasets, which necessitates the development of a data generation framework to reduce the cost of manual annotations.
Although several recent studies have aimed to generate synthetic questions with single-span answers, no study has been conducted on the creation of list questions with multiple, non-contiguous spans as answers.
To address this gap, we propose \ours, an automated framework for generating list QA datasets from unlabeled corpora.
We first convert a passage from Wikipedia or PubMed into a summary and extract named entities from the summarized text as candidate answers.
This allows us to select answers that are semantically correlated in context and is, therefore, suitable for constructing list questions.
We then create questions using an off-the-shelf question generator with the extracted entities and original passage.
Finally, iterative filtering and answer expansion are performed to ensure the accuracy and completeness of the answers.
Using our synthetic data, we significantly improve the performance of the previous best list QA models by exact-match F1 scores of 5.0 on MultiSpanQA, 1.9 on Quoref, and 2.8 averaged across three BioASQ benchmarks.
\end{abstract}

\section{Introduction}
\label{sec:intro}

Extractive question answering (QA) refers to the task of finding answers to questions in the provided text.
Because building a QA system often requires a vast number of human-annotated training examples, recent studies have attempted to reduce annotation costs by generating synthetic datasets from unlabeled corpora~\cite{yang-etal-2017-semi,dhingra-etal-2018-simple,alberti-etal-2019-synthetic,lyu-etal-2021-improving}.
However, these studies have focused only on generating questions with single-span answers and failed to cover list questions that require multiple spans as answers~\cite{voorhees2001overview,tsatsaronis2015overview}.
Although list questions constitute a large portion of the questions asked in practice~\cite{yoon2022sequence}, the automatic generation of list QA datasets has not been sufficiently studied. 

\begin{figure}[t!]
\centering
\begin{subfigure}{.99\columnwidth}
  \centering
  \includegraphics[width=.99\linewidth]{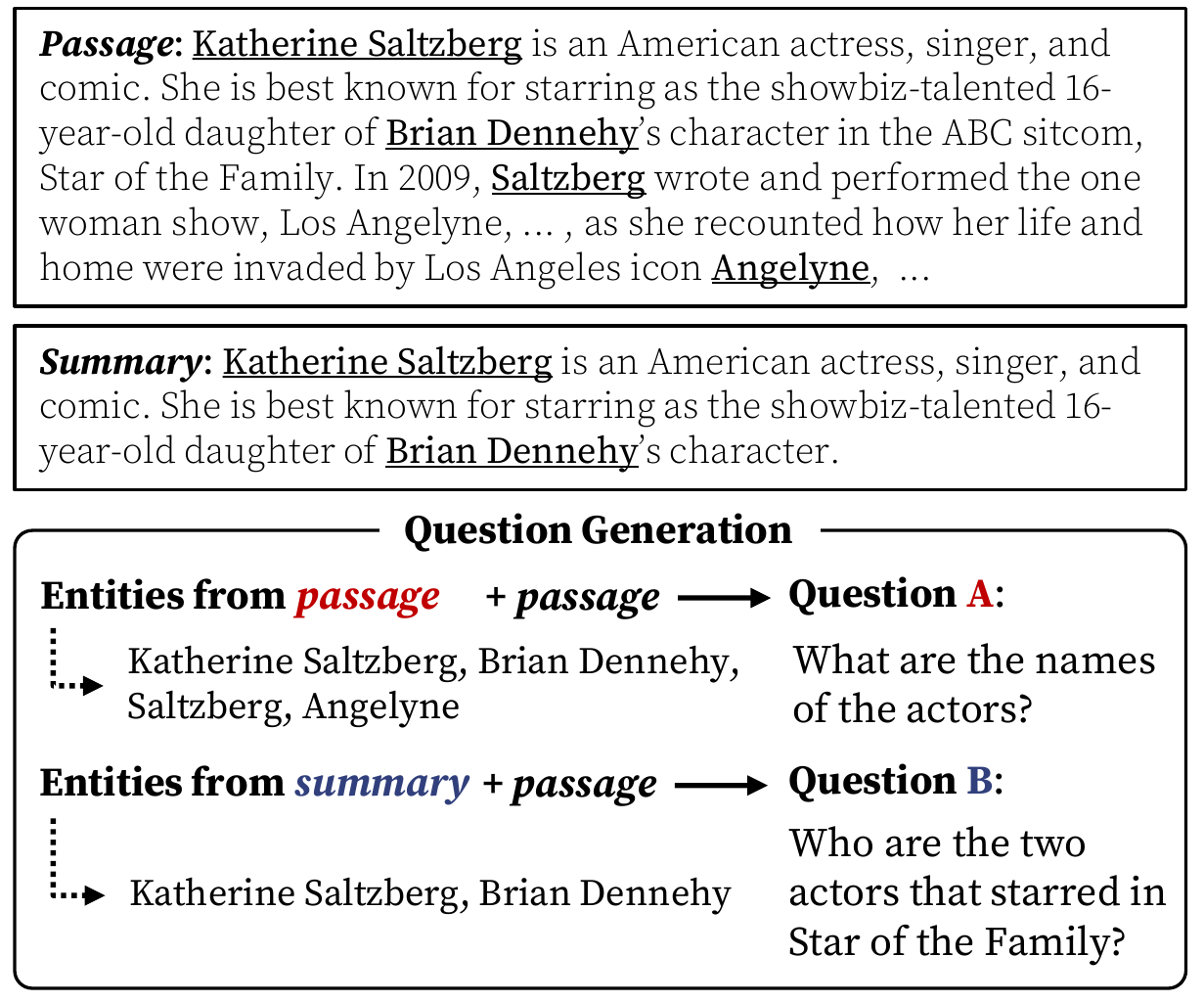} 
\end{subfigure}
\vspace{-3mm}
\caption{
Questions generated conditionally for the same passage with different sets of entities.
When all the named entities in the passage are used, question A about a trivial fact covering all the entities is generated, which may not be effective for training list QA models.
In contrast, using entities from a summary creates a more specific and clearer question, {question B}, because the entities within the summary are usually related by a common topic and fact.
}
\label{fig:motivating}
\vspace{-5mm}
\end{figure}

\begin{figure*}[t!]
\centering
\begin{subfigure}{.99\textwidth}
  \centering
  \includegraphics[width=.975\linewidth]{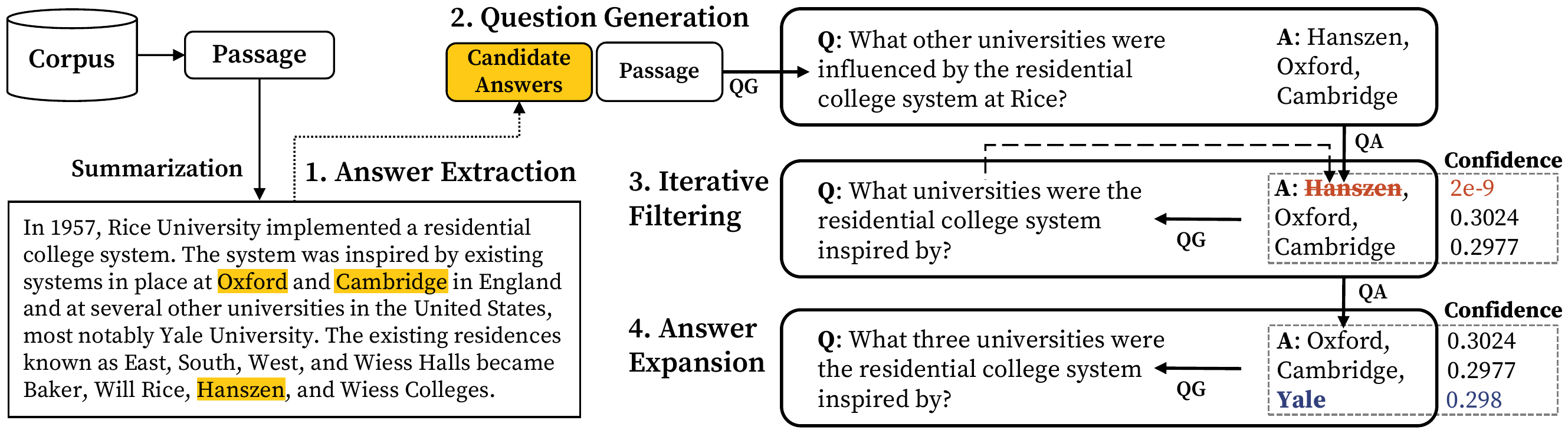} 
\end{subfigure}

\caption{
Overview of \ours. 
(1) Answer extraction: 
the named entities belonging to the same entity type (e.g., organization type) in a summary are extracted by an NER model and used as candidate answers.
(2) Question generation: the candidate answers and the original passage are fed into a QG model to generate list questions.
(3) Iterative filtering: incorrect answers (e.g., Hanszen) are iteratively filtered based on the confidence score assigned by a QA model.
(4) Answer expansion: correct but omitted answers (e.g., Yale) are identified by the QA model.
A threshold value of 0.1 was used in this example.
}
\label{fig:model}
\vspace{-5mm}
\end{figure*}

QA dataset generation frameworks typically consist of answer extraction, question generation, and filtering models that are trained with numerous human-labeled data~\cite{alberti-etal-2019-synthetic,puri-etal-2020-training,shakeri-etal-2020-end,lewis-etal-2021-paq}.
Unfortunately, this supervised approach is not applicable to list QA because existing large QA datasets contain only a few or no list-type questions~\cite{rajpurkar-etal-2016-squad,trischler-etal-2017-newsqa,joshi-etal-2017-triviaqa,rajpurkar-etal-2018-know,yang-etal-2018-hotpotqa,kwiatkowski2019natural}.
Moreover, the automatic generation of list QA datasets presents unique challenges that existing frameworks cannot address. 
First, semantically correlated candidate answers need to be carefully selected because unrelated or excessive candidate answers can result in broad and trivial questions that may not be useful for obtaining list QA models, as depicted in \Cref{fig:motivating}.
In addition, all answers should be accurate and complete, that is, the answer set should not contain incorrect answers or omit correct answers.

In this paper, we present a new framework, \textbf{\ours}, to automatically create \textbf{li}st \textbf{qu}est\textbf{i}on answering \textbf{d}atasets.
Specifically, we first collect passages from unlabeled corpora, such as Wikipedia and PubMed.
Subsequently, a summarization model is used to convert these passages into summaries, and named entity recognition (NER) models extract answers from these summaries.
As entities in the same summary are likely to be semantically correlated to a common topic or fact~\cite{lu2022summarization}, they can be used as suitable candidate answers for list-type questions.
For instance, \Cref{fig:motivating}~illustrates that the entities in the summary (i.e., ``Katherine Saltzberg'' and ``Brian Dennehy'') have a common characteristic in that they appear in the sitcom ``Star of the Family,'' which enables the production of a specific and clear question.
To generate questions, we input the extracted entities and passage into a question generation (QG) model trained on SQuAD~\cite{rajpurkar-etal-2016-squad}.
After the QG process, we improve the quality of the question-answer pairs using a QA model trained on SQuAD.
We perform iterative filtering to ensure accuracy, wherein the QA model is used to eliminate answers with confidence scores lower than a threshold, and the QG model is used to re-generate questions based on the passage and new answer set.
These processes are iterated until the answer set remains unchanged.
For completeness, we perform answer expansion to determine additional spans, which the QA model assigns confidence scores that are higher than the lowest score in the answer set.
The entire process of \ours~allows us to obtain large-scale list QA datasets without relying on hand-labeled list QA data to train each component.

We used five datasets comprising \multispanqa~\cite{li2022multispanqa} and \quoref~\cite{dasigi-etal-2019-quoref} for the general domain, and the \bioasq~7b~\cite{nentidis2019results}, 8b~\cite{nentidis2020overview}, and 9b~\cite{nentidis2021overview} datasets for the biomedical domain.
When the models were trained using our synthetic data, the exact-match F1 scores improved by 5.0 on MultiSpanQA, 1.9 on Quoref, and 2.8 on the three BioASQ datasets, compared with the scores obtained using only human-labeled data.
We conducted extensive ablation studies to confirm the effectiveness of our proposed methods.
In addition, we validated the quality of the generated data and discussed their limitations.

In summary, we made the following contributions:
\begin{itemize}
    \item 
    To the best of our knowledge, this is the first study to introduce a framework for list QA dataset generation.
    We addressed the unique challenges of creating list QA datasets by combining current large-scale models with several methods such as summarization-based answer extraction, iterative filtering, and answer expansion.
    \item
    We significantly improved the performance of the previous best models by an F1 score of 1.9 to 5.0 on five datasets in the general and biomedical domains.
    \item 
    Our code and data have been made publicly available to facilitate further research and real-world applications.\footnote{\url{https://github.com/dmis-lab/LIQUID}}
\end{itemize}

\section{List QA Dataset Generation}
\label{sec:method}

Our goal is to automatically generate list QA data $\mathcal{\tilde{D}}$ from an unlabeled corpus $\mathcal{C}$ to supplement human-labeled data $\mathcal{D}$.
Notably, our framework consists of (1) answer extraction, (2) question generation, (3) iterative filtering, and (4) answer expansion.
The initial data are generated in stages 1 and 2, whereas data refinement is performed in stages 3 and 4.
\Cref{fig:model}~presents the overall process, and \Cref{alg:framework}~(Appendix) details the process using a formal notation.

\subsection{Answer Extraction}
\label{subsec:answer_extraction}
Let $c$ be a passage from corpus $C$.
We first summarize $c$ into $\bar{c}$ and extracted named entities from $\bar{c}$.
Entities of the same types are then used as candidate answers.
Details pertaining to this process are described in the following sections.

\paragraph{Summarization}
One of the most important considerations when selecting candidate answers for list QA is that the answers must be semantically correlated in the given context.
As noted in \Cref{sec:intro}, unrelated candidate answers lead to trivial list questions, because the QG model attempts to construct questions that encompass all given candidate answers.
However, it is difficult to select appropriate candidate answers from all possible spans in a passage.
Instead, we use summarized text (i.e., $\bar{c}$) because a summary is a short snippet that conveys the relevant topics/facts in one or multiple documents; therefore, {similar} phrases (e.g., named entities of the same type) in the same summary are likely to be semantically related to one another and appropriate for list-type questions.
We used the BART$_\texttt{base}$ model~\cite{lewis-etal-2020-bart} trained on the CNN/Daily Mail dataset~\cite{nallapati-etal-2016-abstractive} as the summarization model.

\paragraph{NER} The answers to list questions usually comprise named entities (see \Cref{subsec:answer_distribution}).
In addition, in most cases, the answers to a given question have the same entity type.
Thus, we extract the named entities from a given text and use them as candidate answers. 
We regard entities of the same type as belonging to the same group of answers.
Formally, we obtain $L$ sets of candidate answers $\mathbf{A}_{1},\dots,\mathbf{A}_{L}$ for each summary $\bar{c}$, where $L$ denotes the number of predefined entity types and $\mathbf{A}_{l} = \{a_{l_1},\dots,a_{l_M}\}$ denotes the $l$-th set of candidate answers consisting of $l_M$ entities.
We omit subscript $l$ for simplicity in the following sections.
We used the spaCy NER tagger~\cite{honnibal2020spacy} and BERN2~\cite{sung2022bern2} for the general and biomedical domains, respectively.

\subsection{Question Generation}
\label{subsec:question_generation}

Previous studies attempted to generate questions by training sequence-to-sequence models~\cite{radford2019language,lewis-etal-2020-bart,raffel2020exploring}.
Specifically, in these studies, the QG model considered a single candidate answer $a$ with passage $c$ as an input, which was represented as ``answer: $a$  context: $c$.''
The model was optimized to generate the corresponding question $q$, where the triplet $\langle c, q, a \rangle$ was obtained from large-scale QA datasets.
We adopt this approach with a simple modification to the input format, wherein we concatenate all the candidate answers (i.e., named entities) with commas and prepend them to passage $c$ as follows: ``answer: $a_{1},\dots,a_{M}$ context: $c$.''
The input is then fed into a T5$_\texttt{base}$ model~\cite{raffel2020exploring} trained on a single-span QA dataset, SQuAD.
Interestingly, the model generates appropriate questions, despite not being trained to generate list questions for multiple given answers (see \Cref{subsec:question_analysis}).
After the QG process, we obtain the initial QA instance $\langle c, q, \mathbf{A} \rangle$, where $\mathbf{A}= \{ a_{1},\dots,a_{M} \}$ is the set of $M$ answers.

\subsection{Iterative Filtering}
\label{subsec:filtering}

\paragraph{Filtering}
Because the initial answer set $\mathbf{A}$ can contain incorrect answers to question $q$, we verify the answer set for better accuracy.
Given passage $c$ and question $q$, a single-span QA model is used to obtain the confidence scores for all the candidate answers $a_1,\dots,a_M$.
Answers with confidence scores lower than the threshold $\tau$ are regarded as incorrect and filtered out to yield a refined answer set $\mathbf{A}^{\prime} = \{ a^{\prime}_1,\dots,a^{\prime}_{M^{\prime}} \}$ ($M^{\prime} \le M$).
The triplet $\langle c, q, \mathbf{A}^{\prime} \rangle$ is not used if zero or one answer remains after filtering (i.e., $M^{\prime} \le 1$); otherwise, the instance is passed to the next stage.
For the QA model, we used \robertabase~\cite{liu2019roberta} or \biobertbase~\cite{lee2020biobert} with linear prediction layers for the general and biomedical domains, respectively.
Both models were trained using SQuAD.

\paragraph{Question re-generation}
The initial question $q$ may not perfectly align with the answers obtained after filtering.
Therefore, we re-generate question $q^{\prime}$ based on answer set $\mathbf{A}^{\prime}$ and passage $c$ in the same manner as that described in \Cref{subsec:question_generation}.
Filtering is then performed, and the process is repeated until the current answer set is the same as the previous one or the maximum number of iterations $T$ is reached.

\paragraph{Specifying answer positions}
The start and end positions of the answers are required to train QA models.
Because we extract the answers from summary $\bar{c}$ and use passage $c$ as the evidence text, the correct positions for the answers in the original passage need to be identified.
We address this problem by using the span with the highest confidence score of the QA model as the answer position for an answer string.

\subsection{Answer Expansion}
\label{subsec:answer_exapansion}

The initial set $\mathbf{A}$ is often incomplete, primarily owing to false negatives generated by the NER model.
Although the accuracy of the generated data is improved by excluding incorrect answers during the filtering process, this incompleteness cannot be addressed through filtering.
We address this issue by identifying additional answer spans with confidence scores higher than the lowest confidence in the filtered set $\mathbf{A}^{\prime}$ by using the same QA model as that used for filtering.
A question $q^{\prime\prime}$ based on the expanded set $\mathbf{A}^{\prime\prime}$ is then generated, and triplet $\langle c, q^{\prime\prime}, \mathbf{A}^{\prime\prime} \rangle$ is used as the final instance if the QA model does not filter any answers in $\mathbf{A}^{\prime\prime}$ for question $q^{\prime\prime}$; otherwise, triplet $\langle c, q^{\prime}, \mathbf{A}^{\prime\prime} \rangle$ is used.

\section{Experiments}
\label{sec:experiments}

\begin{table}[t!]
\centering
\footnotesize
\begin{adjustbox}{max width = 0.99\columnwidth}

\begin{tabular}{lrrr}
\toprule
{\textbf{Dataset}} & {\textbf{Train}} & {\textbf{Valid}} & {\textbf{Test}} \\
\midrule
\multicolumn{4}{l}{\textit{General domain}}\\
\midrule
\multispanqa~\cite{li2022multispanqa} & 5,230 & 653 & 653 \\
\quoref~\cite{dasigi-etal-2019-quoref} & 1,766 & 221 & 221 \\
\midrule
\multicolumn{4}{l}{\textit{Biomedical domain}}\\
\midrule
\bioasq~7b~\cite{nentidis2019results} & 556 & 88 & 88 \\
\bioasq~8b~\cite{nentidis2020overview} & 644 & 75 & 75 \\
\bioasq~9b~\cite{nentidis2021overview} & 719 & 94 & 94 \\
\bottomrule
\end{tabular}

\end{adjustbox}

\caption{
Number of questions in list QA benchmark datasets.
``Train,'' ``valid,'' and ``test'' indicate the training, validation, and test sets, respectively.
}
\label{tab:data_statistics}
\vspace{-5mm}
\end{table}

\begin{table*}[t!]
\centering
\footnotesize
\begin{adjustbox}{max width = 0.99\textwidth}

\begin{tabular}{lcccc}
\toprule
\multirow{3}{*}{\textbf{Model}} & \multicolumn{2}{c}{\textbf{\multispanqa}} & \multicolumn{2}{c}{\textbf{\quoref}} \\
\cmidrule(lr){2-3} \cmidrule(lr){4-5} 
\multirow{2}{*}{} & \multicolumn{1}{c}{\textbf{Exact F1 (P/R)}} & \multicolumn{1}{c}{\textbf{Partial F1 (P/R)}} & \multicolumn{1}{c}{\textbf{Exact F1 (P/R)}} & \multicolumn{1}{c}{\textbf{Partial F1 (P/R)}} \\
\midrule
\multicolumn{5}{l}{\textit{\textbf{Baselines}: labeled only ($\mathcal{D}$)}} \\
\midrule
\bertbase~+ Single-span$^{\boldsymbol{\star}}$ & 14.4 (16.2/13.0) & 67.6 (60.3/76.8) & - & - \\
\bertbase~+ Tagger$^{\boldsymbol{\star}}$ & 56.5 (52.5/61.1) & 75.2 (75.9/74.5) & - & - \\
\bertbase~+ Tagger (multi-task)$^{\boldsymbol{\star}}$ & 59.3 (58.1/60.5) & 76.3 (79.6/73.2) & - & - \\
\midrule
\robertabase~+ Single-span & 10.5 (14.4/8.3)\hspace{0.43em} & 63.0 (60.0/66.3) & 55.4 (65.2/48.0) & 69.0 (76.7/62.6) \\
\robertabase~+ Tagger & 62.9 (63.0/62.9) & 78.0 (82.5/73.9) & 81.2 (73.8/90.1) & 85.7 (80.1/92.2) \\
\robertalarge~+ Tagger & {66.4} (62.3/71.2) & \textbf{82.6} (82.1/83.0) & 84.2 (76.1/94.2) & 88.8 (82.6/96.0)\\
\coreflarge~+ Tagger & 64.0 (56.5/73.8) & 81.7 (77.7/86.0) & {86.5} (81.3/92.4) & {89.7} (86.1/93.7)\\
\midrule
\multicolumn{5}{l}{\textit{\textbf{Our models}: synthetic \& labeled ($\mathcal{\tilde{D}}$~$\rightarrow$ $\mathcal{D}$)}} \\
\midrule
\robertabase~+ Single-span & {19.4} (19.7/19.0) & {71.0} (62.9/81.4) & {60.7} (63.8/57.9) & {74.3} (77.4/71.3) \\
\robertabase~+ Tagger & {67.4} (65.7/69.2) & {81.2} (80.9/81.5) & {85.7} (82.3/89.3) & {89.1} (86.5/91.8) \\
\robertalarge~+ Tagger & \textbf{71.4} ({75.0}/68.2) & 80.9 (85.3/77.0) & 86.7 (85.8/87.6) & 90.2 (89.4/91.1)\\
\coreflarge~+ Tagger & 65.8 (64.0/67.8) & 80.2 (79.8/80.5) & \textbf{88.4} (84.8/92.2) & \textbf{91.7} (89.1/94.4) \\
\bottomrule
\end{tabular}
\end{adjustbox}
\caption{
Exact-match and partial-match F1 scores (precision/recall) of QA models on the test sets of \multispanqa~\cite{li2022multispanqa} and \quoref~\cite{dasigi-etal-2019-quoref}.
``Single-span'' and ``tagger'' indicate  single-span extractive and sequence tagging models, respectively.
$\boldsymbol{^\star}$ indicates that the model was implemented by \citet{li2022multispanqa}.
}
\label{tab:results_general}
\vspace{-5mm}
\end{table*}

\subsection{Datasets}
We used (1) {\multispanqa}~\cite{li2022multispanqa}, which consists of English Wikipedia passages with list questions and answers.
The passages and questions were selected from the Natural Questions dataset~\cite{kwiatkowski2019natural}, and the answers were re-annotated by humans to improve their quality.
Note that the model evaluation on the hold-out test set is only available on the official leaderboard.\footnote{https://multi-span.github.io}
We also used another Wikipedia-based dataset, (2) {\quoref}~\cite{dasigi-etal-2019-quoref}, which was originally designed to assess the ability to answer questions that require co-reference reasoning.
The dataset contains some list questions with multiple answers.
We used the original validation set as the test set and split the original training set into the training and validation sets.
For the biomedical domain, we used three datasets provided in the recent (3) {\bioasq} challenges~\cite{tsatsaronis2015overview} comprising \bioasq~7b, 8b, and 9b~\cite{nentidis2019results,nentidis2020overview,nentidis2021overview}.
These datasets comprise evidence texts from biomedical literature with manual annotations by experts.
We sampled data from the training set to form a validation set.
\Cref{tab:data_statistics} presents the dataset statistics.

\subsection{List QA Models}
Because \ours~is a model-agnostic framework, any of list QA models can be used.
We used two types of models: single-span extractive and sequence tagging models (see the paragraphs below).
For the text encoder, we used \roberta~(base and large) and \coreflarge~\cite{ye-etal-2020-coreferential} for the general domain and \biobert~(base and large) for the biomedical domain.
In addition, for MultiSpanQA, we included three BERT-based baseline models~\cite{devlin-etal-2019-bert} used in the study of \citet{li2022multispanqa}.
Among them, the ``multi-task'' model is the previous best model on MultiSpanQA, which is trained with additional objective functions such as span number prediction and structure prediction.
In the fine-tuning stage, we selected the best model checkpoints based on exact-match F1 scores for the validation set at every epoch.
The maximum epochs were set to 20 and 50 for the single-span extractive and sequence tagging models, respectively.
The detailed hyperparameter settings are given in \Cref{appendix:list_qa_models}.

\paragraph{Single-span extractive model}
A conventional approach for performing list QA involves the use of a standard extractive QA model with an absolute threshold, wherein all spans with confidence scores exceeding a threshold are used as the predicted answers, and the threshold is a hyperparameter.

\paragraph{Sequence tagging model}
Single-span extractive models are typically unsuitable for list QA.
In recent studies, the list QA problem has been formulated as a sequence labeling problem, for which models are required to predict the beginning, inside, and outside (i.e., BIO) labels for each token~\cite{segal-etal-2020-simple,yoon2022sequence}.

\begin{table*}[t!]
\centering
\footnotesize
\begin{adjustbox}{max width = 0.99\textwidth}

\begin{tabular}{lcccccc}
\toprule
\multirow{3}{*}{\textbf{Model}} & \multicolumn{2}{c}{\textbf{BioASQ 7b}} & \multicolumn{2}{c}{\textbf{BioASQ 8b}} & \multicolumn{2}{c}{\textbf{BioASQ 9b}} \\
\cmidrule(lr){2-3} \cmidrule(lr){4-5} \cmidrule(lr){6-7}
\multirow{2}{*}{} & \multicolumn{1}{c}{\textbf{Exact F1 (P/R)}} & \multicolumn{1}{c}{\textbf{Partial F1 (P/R)}} & \multicolumn{1}{c}{\textbf{Exact F1 (P/R)}} & \multicolumn{1}{c}{\textbf{Partial F1 (P/R)}} & \multicolumn{1}{c}{\textbf{Exact F1 (P/R)}} & \multicolumn{1}{c}{\textbf{Partial F1 (P/R)}} \\
\midrule
\multicolumn{7}{l}{\textit{\textbf{Baselines}: labeled only ($\mathcal{D}$)}} \\
\midrule
\biobertbase~+ Single-span & 42.1 (55.9/33.8) & 60.2 (82.3/47.5) & 34.4 (44.9/27.9) & 53.5 (40.2/79.9) & 56.1 (46.2/71.3) & 73.8 (70.3/77.7) \\
\biobertbase~+ Tagger & 46.1 (39.7/55.1) & 70.5 (68.5/72.6) & 41.8 (33.5/55.5) & 67.6 (64.0/71.5) & 66.7 (60.1/74.9) & 80.6 (76.4/85.2) \\
\biobertlarge~+ Tagger & {49.5} (40.5/63.6) & {74.6} (70.7/78.9) & {45.0} (34.7/64.0) & {72.2} (65.8/80.0) & {68.2} (60.9/77.5) & {81.4} (76.3/87.2)\\ 
\midrule
\multicolumn{7}{l}{\textit{\textbf{Our models}: synthetic \& labeled ($\mathcal{\tilde{D}}$~$\rightarrow$ $\mathcal{D}$)}} \\
\midrule
\biobertbase~+ Single-span & 51.8 (49.0/55.0) & {70.2} (69.7/70.7) & {44.2} (41.4/47.5) & {65.2} (65.4/65.0) & {64.0} (58.0/71.4) & {76.6} (72.6/81.1) \\
\biobertbase~+ Tagger & {49.0} (41.0/61.0) & {73.1} (70.4/76.0) & {44.2} (36.6/55.8) & {69.4} (67.3/71.7) & {71.5} (67.0/76.6) & {83.2} (80.0/86.7) \\
\biobertlarge~+ Tagger & \textbf{52.3} (44.5/63.5) & \textbf {74.9} (71.9/78.1) & \textbf{46.5} (38.5/58.8) & \textbf{72.3} (68.9/76.1) & \textbf{72.2} (67.3/77.8) & \textbf{83.4} (80.4/86.7) \\ 
\bottomrule
\end{tabular}
\end{adjustbox}
\caption{
Exact-match and partial-match F1 scores (precision/recall) of QA models on the test sets of the \bioasq~7b, 8b, and 9b datasets~\cite{nentidis2019results,nentidis2020overview,nentidis2021overview}.
{Single-span}: single-span extractive model.
{Tagger}: sequence tagging model.}
\label{tab:results_biomedical}
\vspace{-5mm}
\end{table*}

\subsection{Metrics}
\label{subsec:metrics}
Following \citet{li2022multispanqa}, we used strict and relaxed evaluation methods, (1) {exact} and (2) {partial} matches, respectively.
Both methods are based on micro-averaged precision (P), recall (R), and F1 score (F1),\footnote{We use the terms ``exact-match F1 score'' and ``F1 score'' interchangeably.} which are computed as follows:
\begin{equation*}
    \label{equation:metric}
    \begin{aligned}
    & \text{P} = \frac{\sum_{n=1}^{N} \sum_{\hat{a} \in \mathbf{\hat{A}}_{n}} f(\hat{a}, \mathbf{A}_{n}^{*})}
    {\sum_{n=1}^{N} |\mathbf{\hat{A}}_n|}, \\
    & \text{R} = \frac{\sum_{n=1}^{N} \sum_{a^{*} \in \mathbf{A}_{n}^{*}} f(a^{*}, \mathbf{\hat{A}}_{n})}
    {\sum_{n=1}^{N} |\mathbf{A}_n^{*}|},\quad
    \text{F1}= \frac{2 \cdot \text{P} \cdot \text{R}}{(\text{P}+\text{R})}, 
    \end{aligned}
\end{equation*}
where $N$ is the number of questions, $\mathbf{A}_{n}^{*}$ is the set of gold answers, $\mathbf{\hat{A}}_{n}$ is the set of predictions for the $n$-th question, and $|\cdot|$ is the number of elements in the set.
The exact-match and partial-match evaluation methods differ in the scoring function $f$, as detailed in the following paragraphs.

\paragraph{Exact match}
Because the strings of model prediction should ideally be identical to those of the gold answer, the scoring function $f$ is defined as $f(x,\mathbf{Y}) := I_{\mathbf{Y}}(x)$, where $I_{\mathbf{Y}}$ is an indicator function that returns one if $x \in \mathbf{Y}$ and zero; otherwise, $x$ is a span, and $\textbf{Y}$ is a set of spans.

\paragraph{Partial match}
This evaluation considers the overlap between the strings of gold answers and predictions.
The scoring function $f$ is defined as $f(x, \mathbf{Y}) := \texttt{max}_{y \in \mathbf{Y}}(g(x,y)/\texttt{len}(x))$, where $g(x,y)$ is the length of the longest common sub-sequence between $x$ and $y$, and $\texttt{len}(x)$ is the length of string $x$.

\subsection{Results}
\label{subsec:results}

We performed experiments using two setups: (1) {labeled only}, in which the models were trained using only human-labeled data (i.e., the original training data, $\mathcal{D}$), and (2) {synthetic \& labeled}, in which the models were first trained on synthetic data and then fine-tuned with human-labeled data (i.e., $\mathcal{\tilde{D}}$~$\rightarrow$ $\mathcal{D}$).
The validation set was used to determine the best size for the synthetic data based on the F1 score.

Tables~\ref{tab:results_general}~and \ref{tab:results_biomedical}~present the experimental results for the general and biomedical domains, respectively.
The sequence tagging model generally outperformed the single-span extractive model, which is consistent with the results of previous studies~\cite{segal-etal-2020-simple,yoon2022sequence,li2022multispanqa}.
The CorefRoBERTa model showed robust performance on \quoref~because it was designed to capture coreference information.
After the best sequence tagging models for each dataset were fine-tuned (i.e., \robertalarge, \coreflarge, and \biobertlarge~for \multispanqa, \quoref, and \bioasq, respectively), they outperformed their labeled-only counterparts with F1 scores of 5.0 on \multispanqa, 1.9 on \quoref, and 2.8 on the three \bioasq~datasets, respectively.
In particular, our model outperformed the previous best model on MultiSpanQA (i.e., the multi-task model) by an F1 score of 12.1.
For the sequence tagging and single-span extractive models using base-sized encoders, the respective F1 scores improved by 4.5 and 7.1 in the general domain and 3.4 and 9.1 in the biomedical domain, indicating that our framework can be widely utilized with different types of QA models.
While the exact-match F1 scores of large models consistently increased across all datasets, the partial-match F1 scores sometimes decreased because of low recall. 
This could be because the distribution of the number of answers in our synthetic data is relatively skewed to small numbers compared to that in human-labeled data; thus, models with high capacity might have been biased (\Cref{subsec:answer_distribution}).

\section{Analysis}
\label{sec:analysis}

We performed ablation studies of the components and hyperparameters of \ours, as well as data quality comparisons of synthetic and human-labeled data.
We used 140k synthetic question-answer pairs derived from Wikipedia and PubMed.\footnote{We will increase the data size up to 1M for each domain and upload them to our official repository.
}
For human-labeled data, validation sets of MultiSpanQA and \bioasq~9b were used.
Sequence tagging models with \robertabase~and \biobertbase~encoders were used for the list QA model.


\begin{table}[t!]
\centering
\footnotesize
\begin{adjustbox}{max width = 0.99\columnwidth}

\begin{tabular}{lcc}
\toprule
\multirow{1}{*}{\textbf{Model}} & \multicolumn{1}{c}{\textbf{\multispanqa}} & \multicolumn{1}{c}{\textbf{\bioasq~9b}} \\
\midrule
Labeled only & 65.0 & 66.7 \\
\midrule
\multicolumn{3}{l}{\textit{Answer extraction methods}} \\
\midrule
Full Passage  & 70.3 & 67.7 \\
Single Sentence (Passage) & 70.4 & 68.1 \\
Single Sentence (Summary) & 70.7 & 68.4 \\
Full Summary$^\ddagger$ & \textbf{73.0} & \textbf{71.5}  \\
\midrule
\multicolumn{3}{l}{\textit{Number of filtering iterations \& answer expansion}} \\
\midrule
\multicolumn{3}{l}{\textbf{w/o Answer expansion}} \\
$T=0$ (w/o filtering) & 71.2 & 69.1 \\
$T=1$ (filtering once) & 71.3 & 69.9 \\
$T=3$ (iterative filtering) & 71.6 & 70.2 \\ 
\quad {+ Answer expansion$^\ddagger$} & \textbf{73.0} & \textbf{71.5} \\
\bottomrule
\end{tabular}
\end{adjustbox}
\caption{
Ablation study for answer extraction, iterative filtering, and answer expansion.
{$T$}: maximum number of iterations.
{$^\ddagger$}: performance of our final model (\ours).
}
\label{tab:ablation}
\vspace{-5mm}
\end{table}

\subsection{Ablation Study}
\label{subsec:ablation}
\paragraph{Answer extraction methods}

We hypothesized that our summarization-based answer extraction method enabled the selection of semantically correlated candidate answers, which are effective for improving list QA performance.
To validate this hypothesis, we performed experiments with the following three variants for answer extraction while fixing the iterative filtering and answer expansion methods: (1) {full passage}, in which all the named entities in the original passage were used as candidate answers; (2) {single sentence (passage)}, in which named entities from a single sentence in the passage were used as candidate answers; and (3) {single sentence (summary)}, in which the passage was summarized, and the named entities were extracted from a single sentence in the summary.
Note that our final model was ``full summary,'' in which a passage was summarized, and the entities from {all} the sentences in the summary were used.
\Cref{tab:ablation}~indicates that although all three baseline methods improved the performance of the labeled-only models, the performance improvement varied significantly depending on the method.
This demonstrates the importance of selecting an appropriate answer extraction method.
The full-passage model demonstrated the worst performance, because several unrelated named entities were extracted as candidate answers, as mentioned in \Cref{sec:intro}.
The single-sentence (passage) and single-sentence (summary) models exhibited similar performance, and both performed better than the full-passage model because the entities in the same single sentence were more likely to be correlated.
However, using only the candidate answers within a single sentence could misguide the QG model to create simple questions that do not require complex reasoning over multiple sentences, thereby limiting QA performance (see \Cref{subsec:question_analysis}).
Because our method (i.e., full summary) effectively extracted correlated candidate answers from multiple sentences, it significantly outperformed the baselines by F1 scores of 2.3 to 2.7 for the general domain, and 3.1 to 3.8 for the biomedical domain.

\begin{figure}[t!]
\centering

\begin{subfigure}{.9\columnwidth}
  \centering
  \includegraphics[width=.9\linewidth]{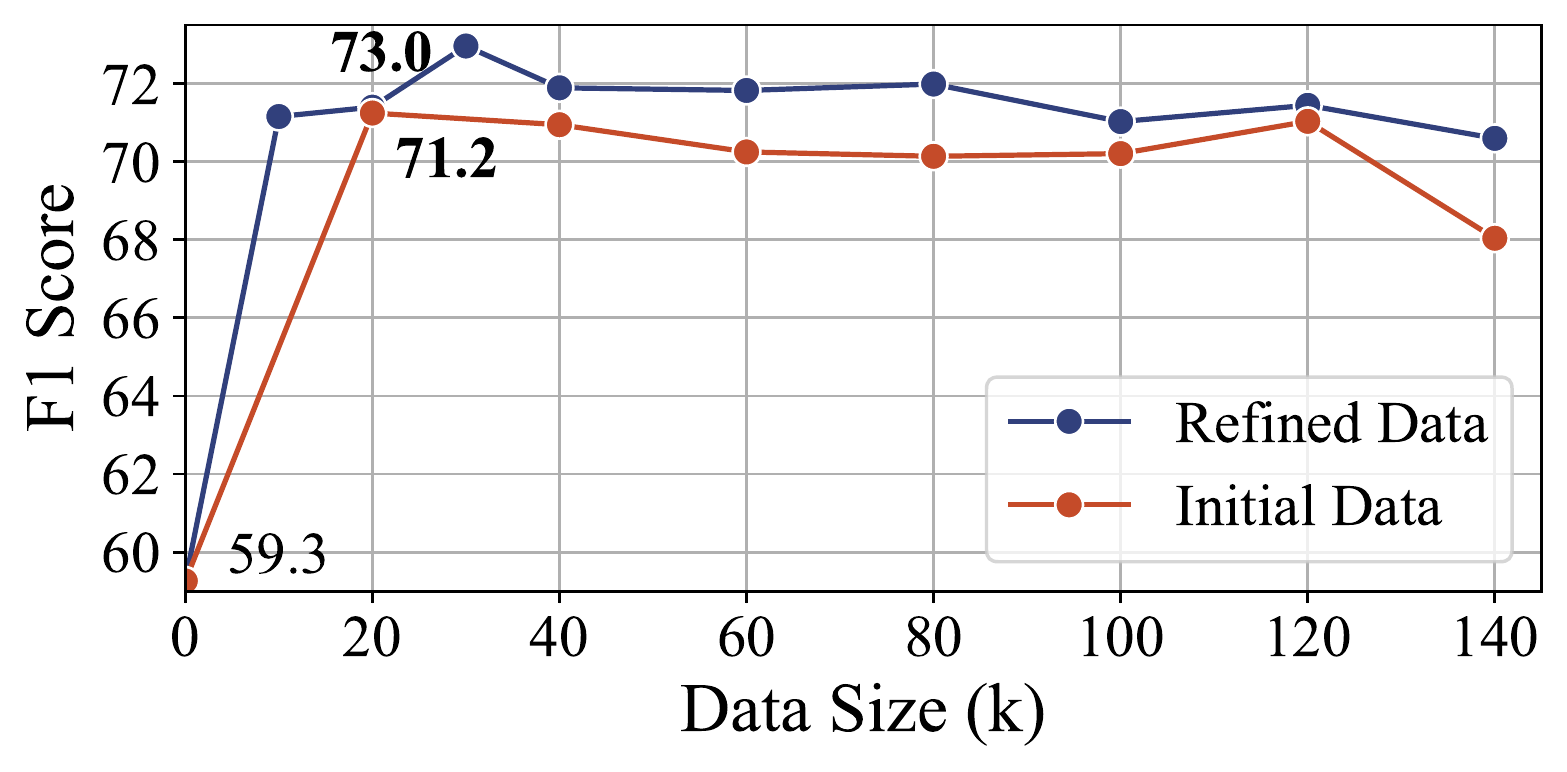} 
  \caption{\multispanqa}
\end{subfigure}

\begin{subfigure}{.9\columnwidth}
  \centering
  \includegraphics[width=.9\linewidth]{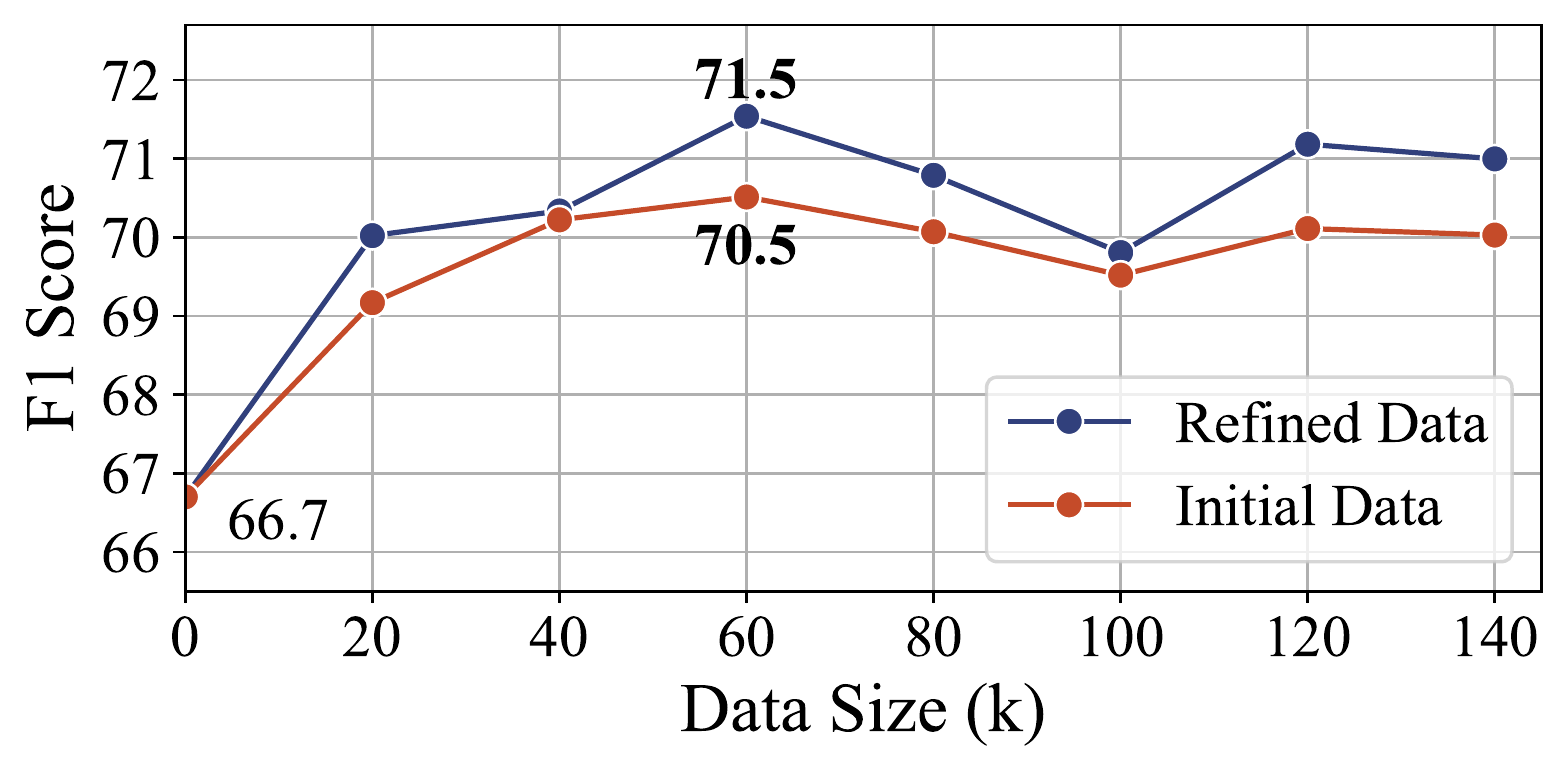}  
  \caption{\bioasq~9b}
\end{subfigure}

\caption{
QA performance for different data sizes of the synthetic data.
``Refined data'' and ``initial data'' represent data generated with and without data refinement (i.e., iterative filtering and answer expansion), respectively.
The scores at $\text{x}=0$ are obtained using only human-labeled data.
}
\label{fig:data_size}
\vspace{-5mm}
\end{figure}

\paragraph{Number of filtering iterations \& answer expansion}
We analyzed the effect of filtering by changing the maximum number of filtering steps $T$.
As presented in \Cref{tab:ablation}, in the absence of filtering (i.e., $T=0$), we obtained F1 scores of 71.2 and 69.1 for the two domains owing to the noise resulting from incorrect answers.
When we used non-iterative filtering (i.e., $T=1$), we obtained better performance with respective F1 scores of 71.3 and 69.9 for the two domains. 
Finally, we achieved the best performance with iterative filtering (i.e., $T=3$) and noted F1 scores of 71.6 and 70.2 for the two domains, respectively.
Using more than three steps was not effective because the filtering process was usually completed before $T=3$.
In addition, when we added the answer expansion method, the respective F1 scores for the two domains further improved by 1.4 and 1.3.

\paragraph{Data size}
We analyzed the variation in the QA performance with the size of the synthetic data for each domain.
\Cref{fig:data_size}~shows that the performance tended to initially increase and then decrease as the data size increased, indicating the existence of an optimal data size.
In addition, the models performed better when iterative filtering and answer expansion methods were used regardless of data size.

\subsection{Answer Distribution}
\label{subsec:answer_distribution}

We analyzed the number of answer spans and answer types in the synthetic data and determined their differences compared to the human-labeled data.
\Cref{appendix:bio_data}~presents a corresponding analysis of the biomedical domain.

\begin{table}[t!]
\centering
\footnotesize
\begin{adjustbox}{max width = 0.99\columnwidth}

\begin{tabular}{lccccl}
\toprule
\textbf{Dataset} & \textbf{2} & \textbf{3} & \textbf{4-5} & \textbf{6-9} & \boldsymbol{$\ge$}\textbf{10 (\%)} \\
\midrule
Synthetic & 76.8 & 18.3 & 4.7 & 0.2 & 0.0   \\
Labeled & 56.8 & 22.7 & 13.9 & 6.1 & 0.5   \\
\bottomrule
\end{tabular}
\end{adjustbox}

\caption{
Number of answer spans in the synthetic and \multispanqa~(i.e., ``labeled'') data.
}
\label{tab:num_of_spans_general}
\end{table}

\paragraph{Number of answer spans}
\Cref{tab:num_of_spans_general}~shows that the number of answer spans in the synthetic data tended to be lower than that in \multispanqa.
The majority (76.8\%) of the questions had two spans, but certain questions (4.9\%) had more than three answers.
We aim to perform further analyses to determine whether dataset bias is caused by limited answer spans.

\begin{table}[t!]
\centering
\footnotesize
\begin{adjustbox}{max width = 0.99\textwidth}

\begin{tabular}{lrr}
\toprule
\textbf{Answer Type} & \textbf{Synthetic} & \textbf{Labeled} \\
\midrule
Person & 39.6\% & 39.1\% \\
GPE/LOC & 29.8\% & 18.6\% \\
ORG & 19.2\% & 2.4\% \\
Numeric & 1.6\% & 3.2\% \\
Others & 9.7\% & 19.4\% \\
Non-entity & 0.0\% & 17.4\% \\
\bottomrule
\end{tabular}
\end{adjustbox}

\caption{
Distribution of answer types for the synthetic and \multispanqa~data.
{GPE/LOC}: (non-)geopolitical regions or locations including countries, cities, mountain ranges, etc.
{ORG}: organizations including companies, institutions, sports teams, etc.
{Others}: all other entities in the world.
{Non-entity}: Any phrase that is not defined as an entity.
}
\label{tab:answer_types_general}
\vspace{-5mm}
\end{table}

\begin{table*}[t!]
\centering
\footnotesize
\begin{adjustbox}{max width = 0.99\textwidth}

\begin{tabular}{lllcc}
\toprule
\multirow{3}{*}{\textbf{\begin{tabular}[t]{@{}p{2.5cm}@{}}Question Type \end{tabular}}} & \multirow{3}{*}{\textbf{Passage \& Answer Spans}} & \multirow{3}{*}{\textbf{Question}} & \multicolumn{2}{c}{\textbf{Percentage}} \\
\cmidrule(lr){4-5}
&  &  & \textbf{Synthetic} & \textbf{Labeled} \\
\midrule
\begin{tabular}[t]{@{}p{2.7cm}@{}}Simple Questions \end{tabular} & \begin{tabular}[t]{@{}p{7.5cm}@{}}  Ya Rab is a 2014 Bollywood movie directed by Hasnain Hyderabadwala starring \textbf{Ajaz Khan}, \textbf{Arjumman Mughal}, \textbf{Raju Kher}, \textbf{Vikram Singh} (actor), \textbf{Imran Hasnee} $\dots$ \end{tabular} & \begin{tabular}[t]{@{}p{3cm}@{}} Who starred in Ya Rab? \end{tabular} & 39.3\% & 26.7\%  \\
\midrule
\begin{tabular}[t]{@{}p{2.7cm}@{}}Lexical Variation\end{tabular} & \begin{tabular}[t]{@{}p{7.5cm}@{}} $\dots$ In June 2007, a Hackday event was hosted at Alexandra Palace by the \textbf{BBC} and \textbf{Yahoo} $\dots$ \end{tabular} & \begin{tabular}[t]{@{}p{3cm}@{}} What \emph{media companies} hosted a Hackday event in 2007? \end{tabular} & 60.7\% & 73.3\% \\
\midrule
\begin{tabular}[t]{@{}p{2.3cm}@{}}Inter-sentence Reasoning\end{tabular} & \begin{tabular}[t]{@{}p{7.5cm}@{}} $\dots$ SBOBET was the shirt sponsor of \textbf{West Ham United}. up until the end of 2012-2013 season. \underline{They} were also the shirt sponsor of \textbf{Cardiff City} for 2010-2011 season $\dots$ \end{tabular} & \begin{tabular}[t]{@{}p{3cm}@{}} What teams did SBOBET sponsor? \end{tabular} & 33.7\% & 57.8\% \\
\midrule
\begin{tabular}[t]{@{}p{2.7cm}@{}}Number of Answers\end{tabular} & \begin{tabular}[t]{@{}p{7.5cm}@{}} $\dots$ While working with her mother, Bundy's uncle offered to pay for her to attend any cookery school in the world. She was accepted into and attended \textbf{Le Cordon Bleu} and \textbf{Le Notre} in Paris, training at Fauchon Patisserie $\dots$ \end{tabular} & \begin{tabular}[t]{@{}p{3cm}@{}} What \emph{two} French cookery schools did Bundy attend? \end{tabular} & 9.0\% & 7.8\% \\
\midrule
\begin{tabular}[t]{@{}p{2.7cm}@{}}Entailment \end{tabular} & \begin{tabular}[t]{@{}p{7.5cm}@{}} $\dots$ Around the same time, \textbf{Zhao Yun} also came to Ye (present-day Handan, Hebei), Yuan Shao's headquarters, where he met \textbf{Liu Bei} again $\dots$ \end{tabular} & \begin{tabular}[t]{@{}p{3cm}@{}} Who were the people who came to Ye? \end{tabular} & 1.1\% & 3.3\% \\
\bottomrule
\end{tabular}
\end{adjustbox}

\caption{
Classification of questions in the synthetic and \multispanqa~data.
All examples are from the synthetic data.
Answers are represented in bold.
See the main text for descriptions of the question types.
}
\label{tab:question_type_general}
\vspace{-5mm}
\end{table*}

\paragraph{Answer types}
We manually classified the types of answers to 100 questions into entity-type categories (\Cref{tab:answer_types_general}).
Notably, both synthetic and labeled datasets consist of many person and geopolitical/location entities.
Apart from these two answer types, humans seem to have a tendency to ask various questions that are not limited to particular answer types. 
This results in relatively few organization-type answers and several others-type answers.
The most notable difference between the two datasets is the number of non-entity answers.
As we relied on NER to extract answers, we could not effectively deal with answers that differed from the named entities defined in the NER system, which is a limitation of our framework.\footnote{The QA model can extract non-entity answers, but we did not find such questions in the 100 sampled examples.}

\subsection{Question Types}
\label{subsec:question_analysis}

We explored the quality and types of list questions generated by the model and asked by humans.
We sampled 100 questions from the synthetic and \multispanqa~data but used only 89 and 90 correct examples for the model- and human-generated questions, respectively, after excluding the incorrect examples (see \Cref{appendix:bio_data} for a corresponding analysis in the biomedical domain). 
We manually classified the questions into the following five categories based on the reasoning required to answer these questions:
\begin{itemize}
    \item \textbf{Simple questions}:
    Questions that were simply derived from evidence texts with few lexical variations.
    \item \textbf{Lexical variation}: 
    Questions that were created with lexical variations using synonyms and hypernyms.
    \item \textbf{Inter-sentence reasoning}:
    Questions that required high-level reasoning such as anaphora, or answers that were distributed across multiple sentences. 
    \item \textbf{Number of answers}:
    Questions that specified the number of answers, which is a characteristic of list questions.
    \item \textbf{Entailment}:
    Questions that required textual entailment based on the evidence texts and commonsense.
\end{itemize}
\Cref{tab:question_type_general}~lists data examples generated by \ours~and the proportion of each question type in the synthetic and labeled data.
It is worth noting that, although the QG model was not tuned for list-type questions, the resulting questions require various types of reasoning to obtain multiple correct answers.
Numerous questions contained lexical variations (60.7\%) while questions that simply matched the evidence texts were avoided.
Some questions involved inter-sentence reasoning (33.7\%) or textual entailment (1.1\%) or specified the number of answer spans (9.0\%).

Additionally, we discovered that most simple questions were generated when the answers belonged to single sentences, indicating the importance of extracting answers spread across multiple sentences to create complex questions.
This also indicates that because the single-sentence models (\Cref{tab:ablation}) extracted candidate answers from single sentences, their performance was relatively low compared to that of \ours, which extracted answers from multiple relevant sentences within the summary.

\subsection{Error Analysis}
We analyzed 411 synthetic data examples and discovered that 50 of them (12.2\%) were wrong.\footnote{In \multispanqa, 10\% of the examples appear to be incorrect, indicating that our framework can generate QA data as accurately as human annotators. Nevertheless, humans are superior in terms of data diversity and quality, as described in Sections \ref{subsec:answer_distribution}~and \ref{subsec:question_analysis}.}
The most dominant error type involved the presence of incorrect answer spans, accounting for 78\% of all errors.
These errors occurred when unrelated entities were grouped into the same set of candidate answers during the answer extraction process, and the filtering model failed to eliminate them.
This indicates that achieving a high accuracy in the answer set still remains a challenge.
In addition, 12\% of the errors can be attributed to missing answers not detected by the expansion model, and 4\% of the errors are incorrect answers added by the model, indicating that the answer expansion method must be improved in future studies.
Finally, the remaining errors (8\%) can be attributed to the QG model, which can be reduced by developing large-scale list QA data or pre-trained models.

\section{Conclusion}
\label{sec:conclusion}
Herein, we introduced \ours, a framework that automatically generates list QA datasets from unlabeled corpora to alleviate the data scarcity problem in this field.
Our synthetic data significantly improved the performance of the current supervised models on five benchmark datasets.
We thoroughly analyzed the effect of each component in \ours~and generated data quantitatively and qualitatively.


\section*{Acknowledgements}
We thank Wonjin Yoon, Gangwoo Kim, and Hajung Kim for the helpful feedback on our manuscript.
We thank Haonan Li for helping us evaluate our models on the MultiSpanQA leaderboard.
This research was supported by (1) National Research Foundation of Korea (NRF-2020R1A2C3010638), (2) the MSIT (Ministry of Science and ICT), Korea, under the ICT Creative Consilience program (IITP-2021-2020-0-01819) supervised by the IITP (Institute for Information \& communications Technology Planning \& Evaluation), and (3) a grant of the Korea Health Technology R\&D Project through the Korea Health Industry Development Institute (KHIDI), funded by the Ministry of Health \& Welfare, Republic of Korea (grant number: HR20C0021).

\bibliography{aaai23}

\clearpage

\appendix

\section{Algorithm}
\label{appendix:algorithm}

\Cref{alg:framework}~describes the data generation process in detail.

\begin{algorithm}[h]
\caption{Data generation process of \ours}
\label{alg:framework}
\textbf{Input}: Source corpus $\mathcal{C}$; Summarization model $\theta_\text{S}$; NER model $\theta_\text{NER}$; QG model $\theta_\text{QG}$; (single-span) QA model $\theta_\text{QA}$; \\
\textbf{Parameter}: The number of passages to sample $K$; The number of entity types $L$; The number of filtering iterations $T$; Threshold for iterative filtering and answer expansion $\tau$; \\
\textbf{Output}: Synthetic QA dataset $\mathcal{\tilde{D}}$;
\begin{algorithmic}[1] 

\STATE $\mathcal{\tilde{D}} \leftarrow \{\}$

\FOR{$k$ $\leftarrow$ $1$ to $K$}

\STATE $c_k \leftarrow \text{RandomSampling}(\mathcal{C})$

\STATE $\bar{c}_k \leftarrow \text{Summarization}(c_k;\theta_\text{S})$

\STATE $\mathbf{A}_1,\dots,\mathbf{A}_L \leftarrow \text{AnswerExtraction}(\bar{c}_k;\theta_\text{NER})$
\STATE \textit{$/*$ Assume that every answer set with less than two elements is excluded from the process.} \hfill $*/$

\FOR{$l$ $\leftarrow$ $1$ to $L$}

\STATE $q_l \leftarrow \text{QuestionGeneration}(\mathbf{A}_l,c_k;\theta_\text{QG})$

\STATE $\mathbf{A}_{l}^{0},\mathbf{A}_{l}^{1},q_{l}^{1} \leftarrow \{\},\mathbf{A}_l,q_{l}$
\STATE $t$ $\leftarrow$ $1$

\WHILE{$\mathbf{A}_{l}^{t-1} \neq \mathbf{A}_{l}^{t}$ \& $t$ $\le$ $T$}
\STATE $\mathbf{A}_{l}^{t+1} \leftarrow \text{Filtering}(c_k, q_l^{t}, \mathbf{A}_{l}^{t};\theta_\text{QA}, \tau)$
\STATE $q_l^{t+1} \leftarrow \text{QuestionGeneration}(\mathbf{A}_{l}^{t+1}, c_k;\theta_\text{QG})$
\STATE $t$ $\leftarrow$ $t + 1$
\ENDWHILE
\STATE $\mathbf{A}_{l}^{\prime},q_{l}^{\prime} \leftarrow \mathbf{A}_{l}^{t-1},q_{l}^{t-1}$

\STATE $\mathbf{A}_{l}^{\prime\prime} \leftarrow \text{AnswerExpansion}(c_k,q_l^{\prime},\mathbf{A}_{l}^{\prime};\theta_\text{QA},\tau)$
\STATE $q_l^{\prime\prime} \leftarrow \text{QuestionGeneration}(\mathbf{A}_{l}^{\prime\prime},c_k;\theta_\text{QG})$
\STATE $\mathbf{\ddot{A}}_{l} \leftarrow \text{Filtering}(c_k, q_l^{\prime\prime}, \mathbf{A}_{l}^{\prime\prime};\theta_\text{QA}, \tau)$

\IF {$\mathbf{A}_{l}^{\prime\prime} = \mathbf{\ddot{A}}_{l}$}
\STATE $\mathcal{\tilde{D}} \leftarrow$ Append($\mathcal{\tilde{D}}$,($c_k,q_l^{\prime\prime},\mathbf{\ddot{A}}_{l}$))
\ELSE
\STATE $\mathcal{\tilde{D}} \leftarrow$ Append($\mathcal{\tilde{D}}$,($c_k,q_l^{\prime},\mathbf{A}_{l}^{\prime\prime}$))
\ENDIF

\ENDFOR

\ENDFOR
\end{algorithmic}
\end{algorithm}

\section{Implementation Details}
\label{appendix:list_qa_models}

\begin{table}[t]
\footnotesize
\centering

\begin{adjustbox}{max width = 0.99\columnwidth}

\begin{tabular}{lcc}
\toprule
\textbf{Model} & \textbf{Data Size} \boldsymbol{$|\mathcal{\tilde{D}}|$} & \textbf{Validation F1} \\
\midrule
\multicolumn{3}{l}{\textit{\multispanqa} / \textit{\quoref}} \\
\midrule
\robertabase~+ Single-span & 60k / 60k & 22.3 / 60.7 \\
\robertabase~+ Tagger & 30k / 90k & 73.0 / 85.7 \\
\robertalarge~+ Tagger & 50k / 60k & 73.6 / 86.7 \\
\coreflarge~+ Tagger & 40k / 20k & 70.3 / 88.4 \\
\midrule
\multicolumn{3}{l}{\textit{\bioasq~7b} / \textit{8b} / \textit{9b}} \\
\midrule
\biobertbase~+ Single-span & 80k / 130k / 80k & 51.8 / 44.2 / 64.0 \\
\biobertbase~+ Tagger & 80k / 80k / 60k & 49.0 / 44.2 / 71.5 \\
\biobertlarge~+ Tagger & 40k / 40k / 60k & 51.5 / 46.3 / 72.2 \\
\bottomrule
\end{tabular}
\end{adjustbox}
\caption{Best synthetic data sizes and corresponding F1 scores on the validation sets for each benchmark dataset.}
\label{tab:hyperparams}
\vspace{-5mm}
\end{table}

\subsection{\ours}
For the unlabeled corpus, we used the 2018-12-20 version of Wikipedia and the 2019-01-02 version of PubMed abstracts for the general and biomedical domains, respectively.
We re-used the trained model parameters available online for the summarization,\footnote{huggingface.co/facebook/bart-large-cnn} QG,\footnote{huggingface.co/mrm8488/t5-base-finetuned-question-generation-ap} and QA models.\footnote{huggingface.co/thatdramebaazguy/roberta-base-squad
}\footnote{huggingface.co/dmis-lab/biobert-base-cased-v1.1-squad}
The minimum and maximum lengths of the output sequence were set to 64 and 128 for the summarization model and 32 and 128 for the QG model.
For the QA model, the maximum lengths of the question and evidence text were set to 128 and 384, respectively, and the trained checkpoints of the \robertabase\footnote{huggingface.co/thatdramebaazguy/roberta-base-squad} and \biobertbase\footnote{huggingface.co/dmis-lab/biobert-base-cased-v1.1-squad} models were used.
We manually selected the thresholds $\tau$, 0.1, and 0.05 for the general and biomedical domains, respectively.
In NER, we excluded the date and species types for the general and biomedical domains, respectively, because they led to trivial candidate answers in our initial experiments.
The best data sizes and best validation F1 scores are listed in \Cref{tab:hyperparams}.

\subsection{List QA models}

We implemented single-span extractive models by modifying the BioBERT-PyTorch repository.\footnote{github.com/dmis-lab/biobert-pytorch}
Sequence tagging models were implemented using the code provided by \citet{yoon2022sequence}.\footnote{github.com/dmis-lab/bioasq-biobert}\footnote{github.com/dmis-lab/SeqTagQA}
For both types of models, we used a batch size of 8 and learning rate of 1e-4.
We searched for the best threshold values of the single-span extractive models using the F1 score on the validation set.
As a result of searching from 0.02 to 0.15 in 0.01 intervals, we selected 0.1 (\quoref), 0.03 (\bioasq~7b), 0.04 (\bioasq~8b), and 0.1 (\bioasq~9b).
For \multispanqa, we used a dynamic threshold search method following \citet{li2022multispanqa}.

\subsection{Datasets and Evaluation}

We obtained \multispanqa,\footnote{multi-span.github.io} \quoref,\footnote{allenai.org/data/quoref} and \bioasq~9b\footnote{participants-area.bioasq.org/datasets} from the official websites.
For \bioasq~7b and 8b, we used document-level evidence texts provided from the repository of \citet{yoon2022sequence}.\footnote{github.com/dmis-lab/SeqTagQA}
We used the evaluation script in the official repository of \citet{li2022multispanqa}.\footnote{github.com/haonan-li/MultiSpanQA}

\section{Biomedical Data Analysis}
\label{appendix:bio_data}

We analyzed synthetic and human-labeled data for the biomedical domain using 140k question-answer pairs generated from PubMed and the \bioasq~9b validation set.

\begin{table}[t!]
\centering
\footnotesize
\begin{adjustbox}{max width = 0.99\columnwidth}

\begin{tabular}{lcccl}
\toprule
\textbf{Dataset} & \textbf{2} & \textbf{3} & \textbf{4-5} & \boldsymbol{$\ge$}\textbf{6 (\%)} \\
\midrule
Synthetic & 61.6 & 24.2 & 13.2 & 1.0 \\ Labeled & 44.3 & 26.5 & 26.5 & 2.8   \\
\bottomrule
\end{tabular}
\end{adjustbox}

\caption{
Number of answer spans in the synthetic and \bioasq~9b data.
}
\label{tab:num_of_spans_bio}
\end{table}

\paragraph{Number of answer spans}
\Cref{tab:num_of_spans_bio}~presents the distribution of the number of answers.
Similar to the results in the general domain (\Cref{tab:num_of_spans_general}), the synthetic data were more skewed toward smaller numbers of answers than the labeled data, but some answers (14.2\%) had four or more spans.

\begin{table}[t!]
\centering
\footnotesize
\begin{adjustbox}{max width = 0.99\textwidth}

\begin{tabular}{lrr}
\toprule
\textbf{Answer Type} & \textbf{Synthetic} & \textbf{Labeled} \\
\midrule
Disease & 47.6\% & 29.6\% \\
Drug/Chemical & 29.5\% & 21.8\% \\
Gene/Protein & 14.2\% & 28.9\% \\
Cell Type/Cell Line & 4.7\% & 0.0\% \\
Others & 0.8\% & 14.8\% \\
Non-entity & 3.1\% & 4.9\% \\
\bottomrule
\end{tabular}
\end{adjustbox}

\caption{
Distribution of answer types for the synthetic and \bioasq~9b data.
}
\label{tab:answer_types_bio}
\vspace{-5mm}
\end{table}

\paragraph{Answer types}
We analyzed the answer types of 100 and 50 examples in the synthetic and labeled data, respectively.
As shown in \Cref{tab:answer_types_bio}, the disease, drug/chemical, and gene/protein types were dominant in both the datasets.
\bioasq~9b does not seem to contain cell types and cell lines and consists of many {others}-type answers, mainly organs.
Unlike the general domain, the synthetic data for the biomedical domain contains non-entity answers, such as descriptions of symptoms and treatments, which are added in the answer expansion stage.

\paragraph{Question types}
We classified 91 and 45 correct examples of 100 and 50 examples for the synthetic and labeled data, respectively, excluding incorrect examples.
Unlike for the general domain (\Cref{tab:question_type_general}), we did not use the {entailment} category and added the {domain knowledge} category, where questions required some biomedical knowledge.
As shown in \Cref{tab:question_type_bio}, many of the synthetic questions contained lexical variations (40.7\%).
The QG model sometimes generates questions that require domain knowledge, but at a much lower rate than the data annotated by human experts.
There are fewer questions with inter-sentence reasoning (8.8\% and 13.3\% for the synthetic and labeled data, respectively) compared to the general-domain corpora (33.7\% and 57.8\% for the synthetic and labeled data, respectively) because (1) the number of selected answers is relatively small and (2) the QG model is not trained with complex questions in the biomedical domain.

\begin{table}[t!]
\centering
\footnotesize
\begin{adjustbox}{max width = 0.99\textwidth}

\begin{tabular}{lrr}
\toprule
\textbf{Question Type} & \textbf{Synthetic} & \textbf{Labeled} \\
\midrule
Simple Question & 58.2\% & 22.2\% \\
Lexical Variation & 40.7\% & 51.1\% \\
Domain Knowledge & 1.1\% & 26.7\% \\
Inter-sentence Reasoning & 8.8\% & 13.3\% \\
Number of Answers & 29.7\% & 15.6\% \\
\bottomrule
\end{tabular}
\end{adjustbox}

\caption{
Classification of questions in the synthetic and \bioasq~9b data.
}
\label{tab:question_type_bio}
\end{table}

\section{Efficiency}
\label{appendix:time_complexity}

We measured the time required to process 10k passages from Wikipedia and PubMed.
We ran our model on an Intel(R) Xeon(R) Silver 4210R CPU @ 2.40GHz and a single 24GB GPU (GeForce RTX 3090).
We used a batch size of eight, that is, eight passages were processed simultaneously.
\Cref{tab:time_complexity}~shows that we can process 10k Wikipedia passages in 72 minutes and 10k PubMed passages in 88 minutes.
For each corpus, we obtained 8,950 and 5,190 initial questions, and 4,274 (47.8\%) and 2,654 (51.1\%) questions after the iterative filtering and answer expansion stages, respectively.
In the question generation, iterative filtering, and answer expansion stages, the processing of passages in Wikipedia took more time than in PubMed because the number of entity types used in the general domain (17 types) was approximately twice than that in the biomedical domain (8 types).
However, the overall process with PubMed was relatively slow mainly because of the run time of the NER model.

\begin{table}[t!]
\centering
\footnotesize
\begin{adjustbox}{max width =\columnwidth}

\begin{tabular}{lrr}
\toprule
\multirow{3}{*}{\textbf{Stage}} & \multicolumn{2}{c}{\textbf{Required Time}} \\ 
\cmidrule(lr){2-3}
& \multicolumn{1}{c}{\textbf{Wikipedia}} & \multicolumn{1}{c}{\textbf{PubMed}} \\
\midrule
1. Answer Extraction \\
\quad - Summarization & 26m 56s & 26m 19s \\
\quad - NER & 1m 43s & 36m \hspace{0.5em}2s \\
\midrule
2. Question Generation & 9m 21s & 4m 39s  \\
\midrule
\begin{tabular}[c]{@{}l@{}}3. Iterative Filtering \& \\\quad Answer Expansion\end{tabular} & 33m 20s & 20m 23s  \\
\midrule
\textbf{Total Time}& 1h 11m 20s & 1h 27m 23s  \\ 
\bottomrule
\end{tabular}
\end{adjustbox}
\caption{
The time taken to process 10k passages.
}
\label{tab:time_complexity}
\vspace{-5mm}
\end{table}

\end{document}